\documentclass[10pt,twocolumn,letterpaper]{article}

\usepackage{cvpr}
\usepackage{multirow}
\usepackage{times}
\usepackage{epsfig}
\usepackage{graphicx}
\usepackage{amsmath}
\usepackage{amssymb}
\usepackage{authblk}

\usepackage[pagebackref=true,breaklinks=true,letterpaper=true,colorlinks,bookmarks=false]{hyperref}

\cvprfinalcopy 

\def\cvprPaperID{9323} 

\ifcvprfinal\fi
\begin{document}

\title{Sketchformer:  Transformer-based Representation for Sketched Structure\vspace{-2mm}}


\author[1]{Leo Sampaio Ferraz Ribeiro\textsuperscript{*}}
\author[2]{Tu Bui\textsuperscript{*}}
\author[2, 3]{John Collomosse}
\author[1]{Moacir Ponti}

\affil[1]{ICMC, Universidade de S{\~a}o Paulo -- S{\~a}o Carlos/SP, Brazil \authorcr
  \tt leo.sampaio.ferraz.ribeiro@gmail.com, ponti@usp.br}
\affil[2]{CVSSP, University of Surrey -- Guildford, Surrey, UK \authorcr
  \{\tt t.bui,j.collomosse\}@surrey.ac.uk}
\affil[3]{Adobe Research, Creative Intelligence Lab --- San Jose, CA, USA}

\maketitle
\renewcommand{\thefootnote}{\fnsymbol{footnote}}
\footnotetext{\textsuperscript{*}These authors contributed equally to this work}

\begin{abstract}
   Sketchformer is a novel transformer-based representation for encoding free-hand sketches input in a vector form, \ie as a sequence of strokes.  Sketchformer effectively addresses multiple tasks: sketch classification, sketch based image retrieval (SBIR), and the reconstruction and interpolation of sketches.  We report several variants exploring continuous and tokenized input representations, and contrast their performance.  Our learned embedding, driven by a dictionary learning tokenization scheme, yields state of the art performance in classification and image retrieval tasks, when compared against  baseline representations driven by LSTM sequence to sequence architectures: SketchRNN and derivatives.    We show that sketch reconstruction and interpolation are improved significantly by the Sketchformer embedding for complex sketches with longer stroke sequences.    
   
\end{abstract}

\section{Introduction}

Sketch representation and interpretation remains an open challenge, particularly for complex and casually constructed drawings.  Yet, the ability to classify, search, and manipulate sketched content remains attractive as gesture and touch interfaces reach ubiquity.  Advances in recurrent network architectures within language processing have recently inspired sequence modeling approaches to sketch (\eg SketchRNN \cite{Eck2018}) that encode sketch as a variable length sequence of strokes, rather than in a rasterized or `pixel' form.  In particular, long-short term memory (LSTM) networks have shown significant promise in learning search embeddings \cite{Xu2018,LS2019} due to their ability to model higher-level structure and temporal order versus convolutional neural networks (CNNs) on rasterized sketches \cite{Bui2017,Qi2016,sketchstyle2017,Hays2016}.  Yet, the limited temporal extent of LSTM restricts the structural complexity of sketches that may be accommodated in sequence embeddings.  In language modeling domain, this shortcoming has been addressed through the emergence of Transformer networks \cite{Devlin2019bert,Dai2019xl,Vaswani2017} in which slot masking enhances the ability to learn longer term temporal structure in the stroke sequence. 

This paper proposes Sketchformer, the first Transformer based network for learning a deep representation for free-hand sketches. We build on the language modeling Transformer architecture of Vaaswani \etal \cite{Vaswani2017} to develop several variants of Sketchformer that process sketch sequences in continuous and tokenized forms.  We evaluate the efficacy of each learned sketch embedding for common sketch interpretation tasks.  We make three core technical contributions:\\
\noindent{\bf1) Sketch Classification.} We show that Sketchformer driven by a dictionary learning tokenization scheme outperforms state of the art sequence embeddings for sketched object recognition over QuickDraw! \cite{QD50}; the largest and most diverse public corpus of sketched objects.\\
\noindent{\bf2) Generative Sketch Model.} We show that for more complex, detailed sketches comprising lengthy stroke sequences, Sketchformer improves generative modeling of sketch -- demonstrated by higher fidelity reconstruction of sketches from the learned embedding.  We also show that for sketches of all complexities, interpolation in the Sketchformer embedding is stable, generating more plausible intermediate sketches for both inter- and intra-class blends.\\
\noindent{\bf3) Sketch based Image Retrieval (SBIR)} We show that Sketchformer can be unified with raster embedding to produce a search embedding for SBIR (after \cite{LS2019} for LSTM) to deliver improved prevision over a large photo corpus (Stock10M).

These enhancements to sketched object understanding, generative modeling and matching demonstrated for a  diverse and complex sketch dataset suggest Transformer as a promising direction for stroke sequence modeling.

\section{Related Work}

Representation learning for sketch has received extensive attention within the domain of visual search. Classical sketch based image retrieval (SBIR) techniques explored spectral, edge-let based, and sparse gradient features the latter building upon the success of dictionary learning based models (e.g. bag of words) \cite{Sivic2003,Bui2015,schneider2014sketch}.  With the advent of deep learning, convolutional neural networks (CNNs) were rapidly adopted to learn search embedding \cite{zhang2016sketchnet}. Triplet loss models are commonly used for visual search in the photographic domain \cite{Wang2014,CTU-ECCV2016,gordo2016deep}, and have been extended to SBIR.  Sangkloy \etal \cite{Hays2016} used a three-branch CNN with triplet loss to learn a general cross-domain embedding for SBIR.  Fine-grained (within-class) SBIR was similarly explored by Yu \etal \cite{Yu2016}.  Qi \etal \cite{Qi2016} instead use contrastive loss to learn correspondence between sketches and pre-extracted edge maps. Bui \etal \cite{BuiArxiv2016,Bui2017} perform cross-category retrieval using a triplet model and combined their technique with a learned model of visual aesthetics \cite{bam} to constrain SBIR using aesthetic cues in \cite{sketchstyle2017}. A quadruplet loss was proposed by \cite{Seddati2017} for fine-grained SBIR. The generalization of sketch embeddings beyond training classes have also been studied \cite{Bui2018,Pang2019}, and parameterized for zero-shot learning \cite{Dey2019}. Such concepts were later applied in sketch-based shape retrieval tasks \cite{xu2018sketch}. Variants of CycleGAN \cite{cyclegan} have also shown to be useful as generative models for sketch \cite{Song2018}. Sketch-A-Net was a seminal work for sketch classification that employed a CNN with large convolutional kernels to accommodate the sparsity of stroke pixels \cite{Yu2016}. Recognition of partial sketches has also been explored  by \cite{seddati2016deepsketch}. Wang \etal \cite{wang2018sketchpointnet} proposed sketch classification by  sampling unordered points of a sketch image to learning a canonical order.

All the above works operate over rasterized sketches \eg converting the captured vector representation of sketch (as a sequence of strokes) to pixel form, discarding temporal order of strokes, and requiring the network to recover higher level spatial structure.  Recent SBIR work has begun to directly input a vector (stroke sequence) representations for sketches \cite{riaz2018learning}, notably SketchRNN; an LSTM based sequence to sequence (seq2seq) variational auto-proposed by Eck \etal \cite{Eck2018}, trained on the largest public sketch corpus `QuickDraw!’ \cite{QD50}. SketchRNN embedding was incorporated in a triplet network by Xu \etal \cite{Xu2018} to search for sketches using sketches. A variation using cascaded attention networks was proposed by \cite{li2018sketchr2cnn} to improve vector sketch classification over Sketch-A-Net. Later, LiveSketch \cite{LS2019} extended SketchRNN to a triplet network to perform SBIR over tens of millions of images, harnessing the sketch embedding to suggest query improvements and guide the user  via relevance feedback.  The limited temporal scope of LSTM based seq2seq models can prevent such representations modeling long, complex sketches, a problem mitigated by our Transformer based model which builds upon the success shown by such architectures for language modeling  \cite{Vaswani2017,Dai2019xl,Devlin2019bert}.  Transformers encode long term temporal dependencies by modeling direct connections between data units. The temporal range of such dependencies was increased via the Transformer-XL \cite{Dai2019xl} and  BERT \cite{Devlin2019bert}, which recently set new state-of-the-art performance on  sentence classification and sentence-pair regression tasks using a cross-encoder. Recent work explores transformer beyond sequence modeling to 2D images \cite{ImageTransformer}.  Our work is first to apply these insights to the problem of sketch modeling, incorporating the Transformer architecture of Vaswani \etal \cite{Vaswani2017} to deliver a multi-purpose embedding that exceeds the state of the art for several common sketch representation tasks.

\section{Sketch Representation}

We propose Sketchformer; a multi-purpose sketch representation from stroke sequence input. In this section we  discuss the pre-processing steps, the adaptions made to the core architecture proposed by Vaswani \etal \cite{Vaswani2017} and the three application tasks.

\subsection{Pre-processing and Tokenization}
Following Eck \etal \cite{Eck2018} we simplify all sketches using the RDP algorithm~\cite{douglas1973algorithms} and normalize stroke length.  Sketches for all our experiments are drawn from QuickDraw50M \cite{QD50} (see Sec.~\ref{sec:exp}; for dataset partitions).  

\noindent{\bf1) Continuous.}  Quickdraw50M sketches are released in the `stroke-3' format where each point $(\delta x, \delta y, p)$ stores its relative position to the previous point together with its binary pen state. To also include the `end of sketch' state, the stroke-5 format is often employed: $(\delta x, \delta y, p_1,p_2,p_3)$, where the the pen states $p_1$ - {\em draw}, $p_2$ - {\em lift} and $p_3$ - {\em end} are mutually exclusive~\cite{Eck2018}. Our experiments with continuous sketch modeling use the `stroke-5' format.

\noindent{\bf2) Dictionary learning.} We build a dictionary of $K$ code words ($K=1000$) to model the relative pen motion \ie $(\delta x, \delta y)$. We randomly sample 100k  sketched pen movements in the training set for clustering via K-means. We allocate 20\% of sketch points for sampling inter-stroke transition, \ie relative transition when the pen is lifted, to balance with the more common within-stroke transitions. Each transition point $(\delta x, \delta y)$ is then assigned to the nearest code word, resulting in a sequence of discrete tokens.  We also include 4 special tokens; a Start of Sketch (SOS) token at the beginning of every sketch, an End of Sketch (EOS) token at the end, a Stroke End Point (SEP) token to be inserted between strokes (indicate pen lifting) and a padding (PAD) token to pad the sketch to a fixed length. 

\noindent{\bf3) Spatial Grid.} The sketch canvas is first quantized into $n \times n$ ($n=100$) square cells, each cell is represented by a token in our dictionary. Given the absolute $(x, y)$ sketch points, we determine which cell contains this point and assign the cell's token to the point. The same four special tokens above are used to complete the sketch sequence.

\begin{figure*}[t!]
    \centering
    \includegraphics[width=0.9\linewidth]{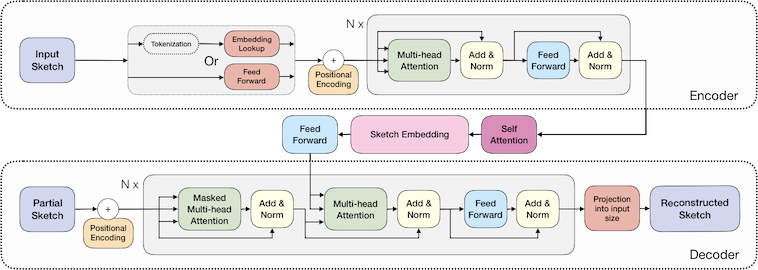}
    \caption{Schematic of the Transformer architecture used for Sketchformer, which utilizes the original architecture of Vaswani \etal \cite{Vaswani2017} but modifies it with an alternate mechanism for formulating the bottleneck (sketch embedding) layer using a self-attention block, as well as configuration changes e.g. MHA head count (see Sec.~\ref{sec:arch}).}
    \label{fig:arch}
\end{figure*}

Fig.~\ref{fig:dict-grid} visualizes sketch reconstruction under the tokenized methods to explore sensitivity to the quantization parameters. Compared with the stroke-5 format (continuous) the tokenization methods are more compact.  Dictionary learned tokenization (Tok-Dict) can have a small dictionary size and is invariant to translation since it is derived from stroke-3.  On the other hand  quantization error could accumulate over longer sketches if dictionary size is too low, shifting the position of strokes closer to the sequence's end. The spatial grid based tokenization method (Tok-Grid), on the other hand, does not accumulate error but is sensitive to translation and yields a larger vocabulary ($n^2)$.

\subsection{Transformer Architecture for Sketch}
\label{sec:arch}
Sketchformer uses the Transformer network of Vaswani \etal~\cite{Vaswani2017}. We add stages (\eg self-attention and modified bottleneck) and adapt parameters in their design to learn a multi-purpose representation for stroke sequences, rather than language.  A transformer network consists of an encoder and decoder blocks, each comprising several layers of multihead attention followed by a feed forward network. Fig.~\ref{fig:arch} illustrates the architecture with dotted lines indicating re-use of architecture stages from \cite{Vaswani2017}. In Fig.~\ref{fig:all-tasks-diagram} we show how our learned embedding is used across multiple applications. Compared to \cite{Vaswani2017} we use 4 MHA blocks versus 6 and a feed-forward dimension of 512 instead of 2048. Unlike traditional sequence modeling methods (RNN/LSTM) which learns the temporal order of current time steps from previous steps (or future steps in bidirectional encoding), the attention mechanism in transformers allows the network to decide which time steps to focus on to improve the task at hand. Each multihead attention (MHA) layer is formulated as such:
\begin{figure}[b!]
    \centering
     \includegraphics[width=1.0\linewidth]{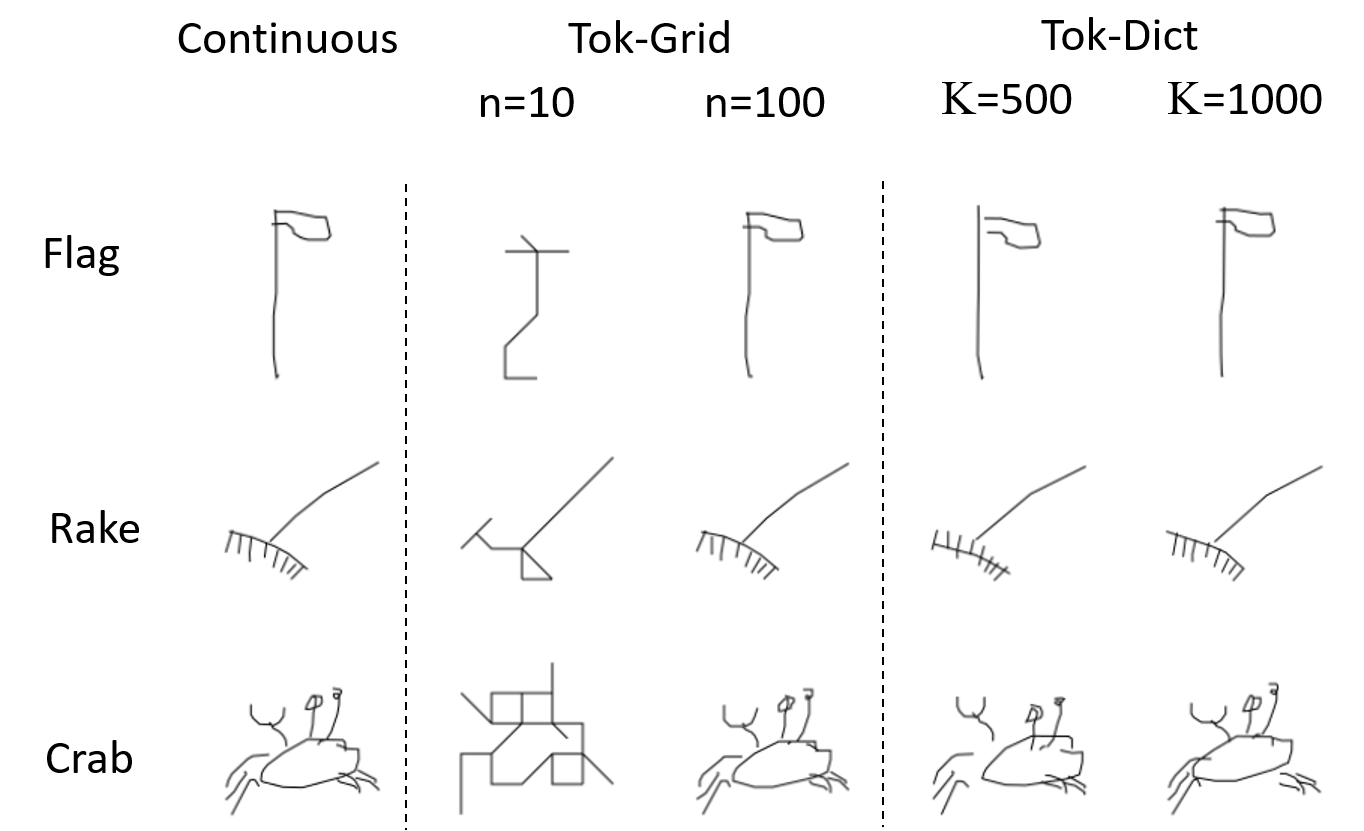}

    \caption{Visualizing the impact of quantization on the reconstruction of short, median and long sequence length sketches.  Grid sizes of $n=[10,100]$ (Tok-Grid) and dictionary sizes of $K=[500,1000]$ (Tok-Dict). Sketches are generated from the tokenized sketch representations, independent of transformer.}
    \label{fig:dict-grid}
\end{figure}\begin{align}
    SHA(k, q, v) &= softmax(\alpha qk^T)v\\
    MHA(k, q, v) &= [SHA_0(kW^k_0,qW^q_0,vW^v_0), ...\\
    & SHA_m(kW^k_m,qW^q_m,vW^v_m)]W^0
\end{align}
where $k$, $q$ and $v$ are respective \textit{Key}, \textit{Query} and Value inputs to the single head attention (SHA) module. This module computes the similarity between pairs of \textit{Query} and \textit{Key} features, normalizes those scores and finally uses them as a projection matrix for the \textit{Value} features. The multihead attention (MHA) module concatenates the output of multiple single heads and projects the result to a lower dimension. $\alpha$ is a scaling constant and $W^{(.)}_{(.)}$ are learnable weight matrices.

\begin{figure*}[t!]
    \centering
    \includegraphics[width=0.9\linewidth]{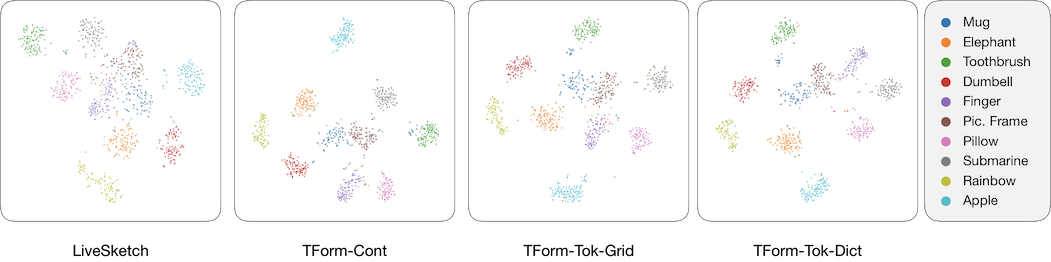}
    \caption{t-SNE visualization of the learned embedding space from Sketchformer's three variants, compared to LiveSketch (left) and computed over the QD-862k test set; 10 categories and 1000 samples were randomly selected.}
    \label{fig:tsne}
\end{figure*}
The MHA output is fed to a positional feed forward network (FFN), which consists of two fully connected layers with ReLU activation. The MHA-FFN ($F(.)$) blocks are the basis of the encoder side of our network ($\mathcal{E(.)}$):
\begin{align}
    FFN(x) &= max(0, xW^f_1+b^f_1)W^f_2 + b^f_2\\
    F(x) &= \overline{FFN}(\overline{MHA}(x,x,x)) \\
    \mathcal{E}(x) &= F_N(..(F_1(x)))
\end{align}
where $\overline{X}$ indicates layer normalization over X and N is number of the MHA-FFN units $F(.)$.

The decoder takes as inputs the encoder output and target sequence in an auto-regressive fashion. In our case we are learning an transformer autoencoder so the target sequence is also the input sequence shifted forward by 1:
\begin{align}
    G(h, x) &= \overline{FFN}(\overline{MHA}_2(h, h, \overline{MHA}_1(x,x,x)))\\
    \mathcal{D}(h, x*) &= G_N(h, G_{N-1}(h, ...G_1(h, x*)))
\end{align}
where $h$ is the encoder output, $x*$ is the shifted auto-regressive version of input sequence $x$.


The conventional transformer is designed for language translation and thus does not provide a feature embedding as required in Sketchformer (output of $\mathcal{E}$ is also a sequence of vectors of the same length as $x$). To learn a compact representation for sketch we propose to apply self-attention on the encoder output, inspired by \cite{sukhbaatar2015end}:
\begin{align}
    s &= softmax(tanh(hK^T+b)v)\\
    z &= \sum_i{s_i h_i}
\end{align}
which is similar to SHA however the Key matrix $K$, Value vector $v$ and bias $b$ are now trainable parameters. This self-attention layer learns a weight vector $s$ describing the importance of each time step in sequence $h$, which is then accumulated to derive the compact embedding $z$. On the decoder side, $z$ is passed through a FFN to resume the original shape of $h$.  These are the key novel modifications to the original Transformer architecture of Vaswani \etal (beyond above-mentioned parameter changes). 

We also had to change how masking worked on the decoder. The Transformer uses a padding mask to stop attention blocks from giving weight to out-of-sequence points. Since we want a meaningful embedding for reconstruction and interpolation, we removed this mask from the decoder, forcing our transformer to learn reconstruction without previously knowing the sequence length and using only the embedding representation.


\subsubsection{Training Losses}
We employ two losses in training Sketchformer. A classification (softmax) loss is connected to the sketch embedding $z$ to preserve semantic information while a reconstruction loss ensures the decoder can reconstruct the input sequence from its embedding. If the input sequence is continuous (\ie stroke-5) the reconstruction loss consists of a $L^2$ loss term modeling relative transitions $(\delta x, \delta y)$ and a 3-way classification term modeling the pen states. Otherwise the reconstruction loss uses softmax to regularize a dictionary of sketch tokens as per a language model. We found these losses simple yet effective in learning a robust sketch embedding.  Fig.~\ref{fig:tsne} visualizes the learned embedding for each of the three pre-processing variants, alongside that of a state of the art sketch encoding model using stroke sequences \cite{LS2019}.

\subsection{Cross-modal Search Embedding}
\label{sec:search}

To use our learned embedding for SBIR, we follow the joint embedding approach first presented in \cite{LS2019} and train an auxiliary network that unifies the vector (sketch) and raster (image corpus) representations into a common subspace. 

\begin{figure}
    \centering
    \includegraphics[width=0.99\linewidth,height=7cm]{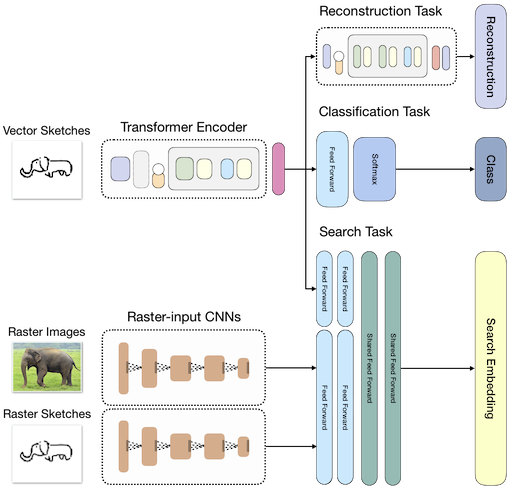}
    \caption{Schematic showing how the learned sketch embedding is leveraged for sketch synthesis (reconstruction/interpolation), classification and cross-modal retrieval experiments (see encoder/embedding inset, refer to Fig.~\ref{fig:arch} for detail).  Classification appends fully-connected (fc) and softmax layers to the embedding space.  Retrieval tasks require unification with a raster (CNN) embedding for images \cite{Bui2018} via several fc layers trained via triplet loss.}
    \label{fig:all-tasks-diagram}
\end{figure}

This auxiliary network is composed of four fully connected layers (see Fig.~\ref{fig:all-tasks-diagram}) with ReLU activations. These are trained within a triplet framework and have input from three pre-trained branches: an anchor branch that models vector representations (our Sketchformer), plus positive and negative branches extracting representations from raster space.

The first two fully connected layers are domain-specific and we call each set $F_V(.), F_R(.)$, referring to vector-specific and raster-specific. The final two layers are shared between domains; we refer to this set as $F_S(.)$. Thus the end-to-end mapping from vector sketch and raster sketch/image to the joint embedding is:
\begin{align}
    u_v &= F_S(F_V(\mathcal{E}(x_v)))\\
    u_r &= F_S(F_R(\mathcal{P}(x_r)))
\end{align}
where $x_v$ and $x_r$ are the input vector sketches and raster images respectively, and $u_v$ and $u_r$ their corresponding representations in the common embedding. $\mathcal{E}(.)$ is the network that models vector representations and $\mathcal{P}(.)$ is the one for raster images. In the original LiveSketch~\cite{LS2019}, $\mathcal{E}(.)$ is a SketchRNN~\cite{Eck2018}-based model, while we employ our multi-task Sketchformer encoder instead. For $\mathcal{P}(.)$ we use the same off-the-shelf GoogLeNet-based network, pre-trained on a joint embedding task (from~\cite{Bui2018}).

The training is performed using triplet loss regularized with the help of a classification task. Training requires an aligned sketch and image dataset \ie a sketch set and image set that share the same category list. This is not the case for Quickdraw, which is a sketch-only dataset without a corresponding image set. Again following~\cite{LS2019}, we use the raster sketch as a medium to bridge vector sketch with raster image. The off-the-shelf $\mathcal{P}(.)$ (from \cite{Bui2018}) was trained to produce a joint embedding model unifying raster sketch and raster image; This allowed the authors train the $F_R$, $F_V$ and $F_S$ sets using vector and raster versions of sketch only. By following the same procedure, we eliminate the need of having an aligned image set for Quickdraw as our network never sees an image feature during training.

The training is implemented in two phases. At phase one, the anchor and positive samples are vector and raster forms of random sketches in the same category while raster input of the negative branch is sampled from a different category. At phase two, we sample hard negatives from the same category with the anchor vector sketch and choose the raster form of the exact instance of the anchor sketch for the positive branch. The triplet loss maintains a margin between the anchor-positive and anchor-negative distances:
 \begin{equation}
 \small
 \begin{split}
     \mathcal{L}_T(x, x_+,x_-) &= max(0, \left|F_S(F_V(\mathcal{E}(x)))- F_S(F_R(\mathcal{P}(x_+)))\right|\\
     &- \left|\left|F_S(F_V(\mathcal{E}(x)))- F_S(F_R(\mathcal{P}(x_-)))\right|\right| + m) 
 \end{split}
 \end{equation}
and margin $m=0.2$ in phase one, $m=0.05$ in phase two.

\section{Experiments and Discussion}
\label{sec:exp}
We evaluate the performance of the proposed transformer embeddings for three common tasks; sketch classification, sketch reconstruction and interpolation, and sketch based image retrieval (SBIR).  We compare against two baseline sketch embeddings for encoding stroke sequences; SketchRNN \cite{Eck2018} (also used for search in \cite{Xu2018}) and LiveSketch \cite{LS2019}. We evaluate using sketches from  QuickDraw50M \cite{QD50}, and a large corpus of photos (Stock10M).

\begin{table}[b!]
    \centering
        \begin{tabular}{lll}
            \ttfamily Method & \ttfamily ~~ & \ttfamily mAP\% \\                \hline
            \hline
            \multirow{2}{*}{Baseline} & \multicolumn{1}{l}{LiveSketch \cite{LS2019}} & \multicolumn{1}{l}{72.93} \\\cline{2-3}
                                      & \multicolumn{1}{l}{SketchRNN \cite{Eck2018}} & \multicolumn{1}{l}{67.69} \\\hline
            \multirow{3}{*}{Shuffled} & \multicolumn{1}{l}{TForm-Cont} & \multicolumn{1}{l}{76.95} \\\cline{2-3}
                                     & \multicolumn{1}{l}{TForm-Tok-Grid} & \multicolumn{1}{l}{76.22} \\\cline{2-3}
                                     & \multicolumn{1}{l}{TForm-Tok-Dict} & \multicolumn{1}{l}{76.66} \\\hline
            \multirow{3}{*}{Proposed} & \multicolumn{1}{l}{TForm-Cont} & \multicolumn{1}{l}{77.68} \\\cline{2-3}
                                    & \multicolumn{1}{l}{TForm-Tok-Grid} & \multicolumn{1}{l}{77.36} \\\cline{2-3}
                                    & \multicolumn{1}{l}{TForm-Tok-Dict} & \multicolumn{1}{l}{{\bf 78.34}} \\\hline
        \end{tabular}

    \caption{Sketch classification results over QuickDraw! \cite{QD50} for three variants of the proposed transformer embedding, contrasting each to models learned from randomly permuted stroke order. Comparing to two recent LSTM based approaches for sketch sequence encoding \cite{LS2019, Eck2018}.}
    \label{tab:res_class}
\end{table}

{\bf QuickDraw50M}  \cite{QD50} comprises over 50M sketches of 345 object categories, crowd-sourced within a gamified context that encouraged casual sketches drawn at speed.  Sketches are often messy and complex in their structure, consistent with tasks such as SBIR. Quickdraw50M captures sketches as stroke sequences, in contrast to earlier raster-based and 
less category-diverse datasets such as TUBerlin/Sketchy. We sample 2.5M sketches randomly with even class distribution from the public Quickdraw50M training partition to create training set (QD-2.5M) and use the public test partition of QuickDraw50M (QD-862k) comprising 2.5k $\times 345 = 862k$ sketches to evaluate our trained models. For SBIR and interpolation experiments we sort QD-862k by sequence length, and sample three datasets (QD345-S, QD345-M, QD345-L) at centiles 10, 50 and 90 respectively to create a set of short, medium and long stroke sequences.  Each of these three datasets samples one sketch per class at random from the centile yielding three evaluation sets of 345 sketches.  We sampled an additional query set QD345-Q for use in sketch search experiments, using the same 345 sketches  as LiveSketch \cite{LS2019}.  The median stroke lengths of QD345-S, QD345-M, QD345-L are 30, 47 and 75 strokes respectively (after simplification via RDP \cite{douglas1973algorithms}).

{\bf Stock67M} is a diverse, unannotated corpus of photos
used in prior SBIR work \cite{LS2019} to evaluate large-scale SBIR retrieval performance.  We sample 10M of these images at random for our search corpus (Stock10M).


\subsection{Evaluating Sketch Classification}

We evaluate the class discrimination of the proposed sketch embedding via attaching dense and softmax layers to the transformer encoder stage, and training a 345-way classifier on QD2.5M.   Table~\ref{tab:res_class} reports the classification performance over QD-862k for each of the three proposed transformer embeddings, alongside two LSTM baselines --  the SketchRNN \cite{Eck2018} and LiveSketch \cite{LS2019} variational autoencoder networks.  Whilst all transformers outperform the baseline, the tokenized variant of the transformer based on dictionary learning (TForm-Tok-Dict) yields highest accuracy.  We explore this further by shuffling the order of the sketch
strokes retraining the transformer models from scratch. We were surprised
to see comparable performance, suggesting this
gain is due to spatial continuity rather than temporal information.

\begin{figure}[t!]

    \includegraphics[width=1.0\linewidth]{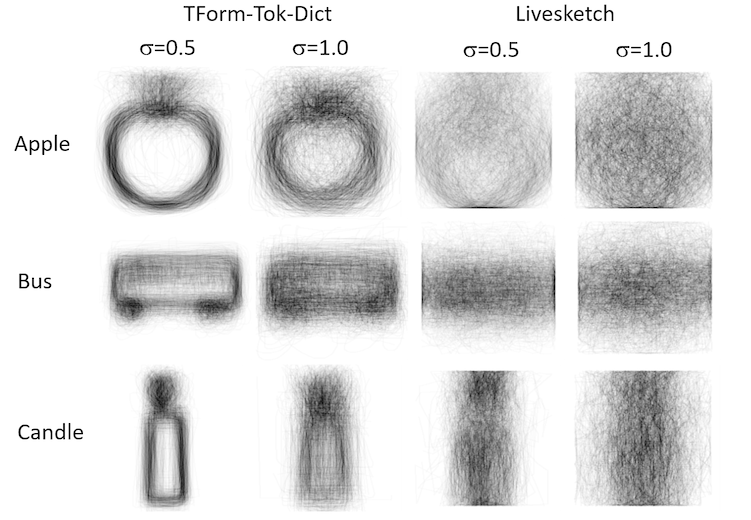}
\caption{Visualization of sketches reconstructed from mean embedding for 3 object categories. We add  
    Gaussian noise with standard deviation $\sigma=0.5$ and $\sigma=1.0$ to the mean embedding of three example categories on the Quickdraw test set. The reconstructed sketches of Tform-Tok-Dict retain salient features even with high noise perturbation.}
    \label{fig:noise}
\end{figure}

\subsection{Reconstruction and Interpolation}

We explore the generative power of the proposed embedding by measuring the degree of fidelity with which: 1) encoded sketches can be reconstructed via the decoder to resemble the input; 2) a pair of sketches may be interpolated within, and synthesized from, the embedding.  The experiments are repeated for short (QD345-S), medium (QD345-M) and long (QD345-L) sketch complexities.  We assess the fidelity of sketch reconstruction and the visual plausibility of interpolations via Amazon Mechanical Turk (MTurk).  MTurk workers are presented with a set of reconstructions or interpolations and asked to make a 6-way preference choice; 5 methods and a 'cannot determine' option.  Each task is presented to five unique workers, and we only include results for which there is $>50\%$ (\ie $>2$ worker) consensus on the choice.

\begin{figure}[t!]
    \centering
    \includegraphics[width=0.9\linewidth]{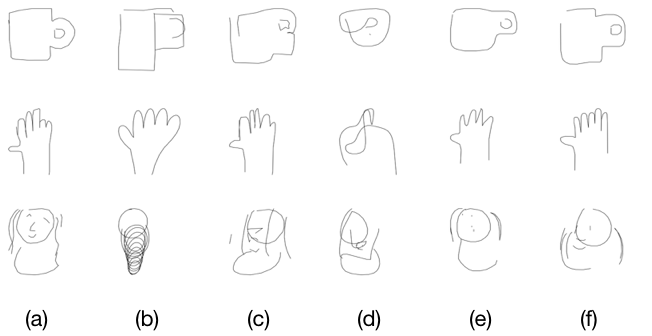}
    \caption{Representative sketch reconstructions from each of the five embeddings evaluated in Table~\ref{tab:res_rec}. (a) Original, (b) SketchRNN, (c) LiveSketch, (d) TForm-Cont, (e) TForm-Tok-Grid and (f) TForm-Tok-Dict. The last row represents a hard-to-reconstruct sketch.}
    \label{fig:recons}
\end{figure}
\begin{table}[t!]
    \centering
        \begin{tabular}{llccc}
            \ttfamily Method & \ttfamily ~~ & \ttfamily Short & \ttfamily Mid & \ttfamily Long \\                \hline
            \hline
            \multirow{2}{*}{Baseline} & \multicolumn{1}{l}{LiveSketch \cite{LS2019}} & \multicolumn{1}{l}{\bf 62.1} & \multicolumn{1}{l}{\bf 59.2} & \multicolumn{1}{l}{ 27.8}\\\cline{2-5}
                                      & \multicolumn{1}{l}{SketchRNN \cite{Eck2018}} & \multicolumn{1}{l}{4.05} & \multicolumn{1}{l}{3.70} & \multicolumn{1}{l}{1.38} \\\hline
            \multirow{3}{*}{Proposed} & \multicolumn{1}{l}{TForm-Cont} & \multicolumn{1}{l}{0.00} & \multicolumn{1}{l}{0.00} & \multicolumn{1}{l}{0.00}\\\cline{2-5}
                                    & \multicolumn{1}{l}{TForm-Tok-Grid} & \multicolumn{1}{l}{6.75} & \multicolumn{1}{l}{6.10} & \multicolumn{1}{l}{5.56}\\\cline{2-5}
                                    & \multicolumn{1}{l}{TForm-Tok-Dict} & \multicolumn{1}{l}{{24.3}} & \multicolumn{1}{l}{28.4} & \multicolumn{1}{l}{\bf 51.4}\\\hline
            \multirow{1}{*}{~} & \multicolumn{1}{l}{{\em Uncertain}} & \multicolumn{1}{l}{2.70} & \multicolumn{1}{l}{2.47} & \multicolumn{1}{l}{13.9}\\\hline        \end{tabular}

    \caption{User study quantifying accuracy of sketch reconstruction.  Preference is expressed  by 5 independent workers, and results with $>50\%$ agreement are included.  Experiment repeated for short, medium and longer stroke sequences. For longer sketches, the proposed transformer method TForm-Tok-Dict is preferred.}
    \label{tab:res_rec}
\end{table}

\begin{figure*} [t!]
    \centering
            \includegraphics[width=1.0\linewidth]{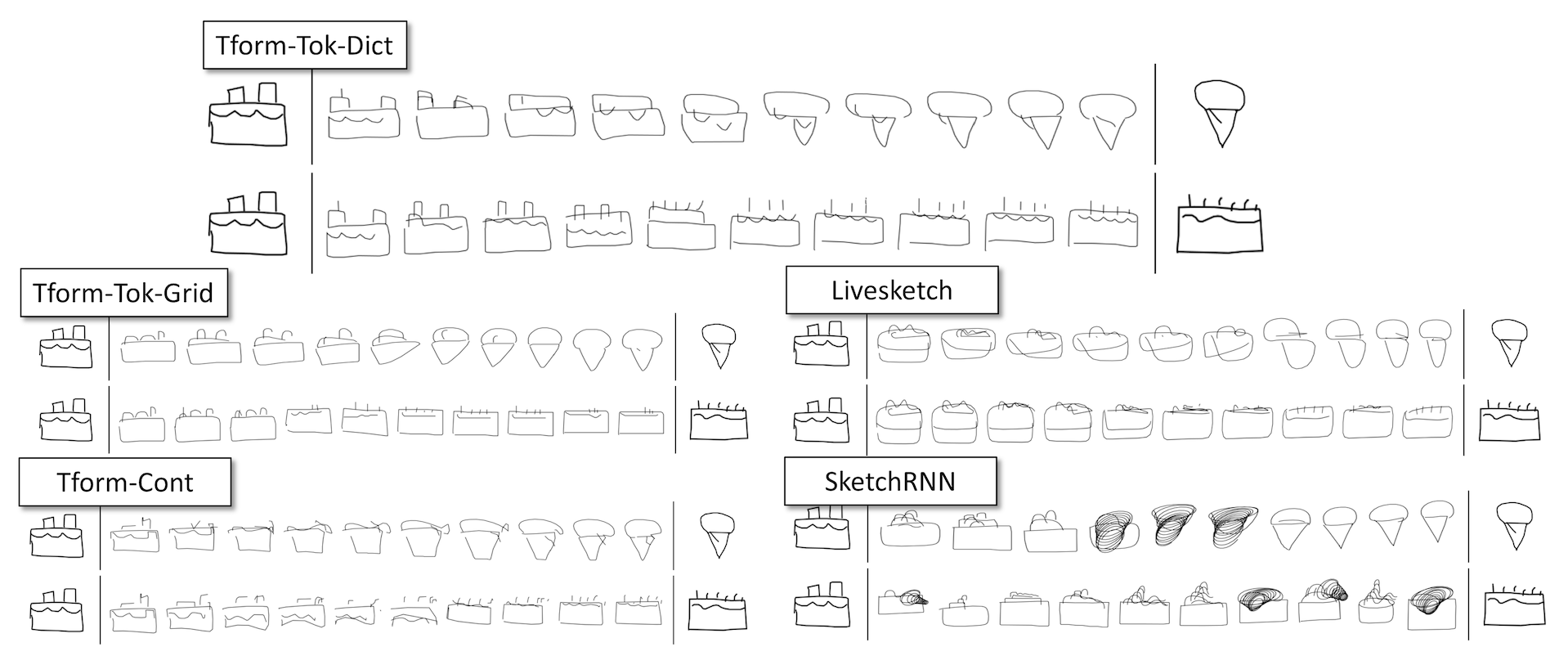}
    \caption{Representative sketch interpolations from each of the five embeddings evaluated in Table~\ref{tab:res_interp}. For each embedding: (first row) inter-class interpolation from 'birthday cake' to `ice-cream' and (second row) intra-class interpolation between two `birthday cakes'.}
    \label{fig:interpfig}
\end{figure*}
Reconstruction results are shown in Table~\ref{tab:res_rec} and favor the LiveSketch \cite{LS2019} embedding for short or medium length strokes, with the proposed tokenized transformer (TForm-Tok-Dict) producing better results for more complex sketches aided by the improved representational power of transformer for longer stroke sequences. Fig~\ref{fig:recons} provides representative visual examples for each sketch complexity.

We explore interpolation in Table~\ref{tab:res_interp} blending between pairs of sketches within (intra-) class and between (inter-) class.  In all cases we encode sketches separately to the embedding, interpolate via slerp (after \cite{Eck2018,LS2019} in which slerp was shown to offer best performance), and decode the interpolated point to generate the output sketch.  Fig.~\ref{fig:interpfig} provides visual examples of inter- and intra- class interpolation for each method evaluated.  In all cases the proposed tokenized transformer (TForm-Tok-Dict) outperforms other transformer variants and baselines, although the performance separation is narrower for shorter strokes echoing results of the reconstruction experiment.  The stability of our representation is further demonstrated via local sampling within the embedding in Fig.~\ref{fig:noise}.

\begin{table}[t!]
    \centering
    \small
        \begin{tabular}{llcc}
            \ttfamily Partition & \ttfamily Embedding & \ttfamily Intra- & \ttfamily Inter- \\                \hline
            \hline

            \multirow{5}{*}{Short ($10^\mathrm{th}$ cent.)}  & \multicolumn{1}{l}{SketchRNN \cite{Eck2018}} & \multicolumn{1}{l}{0.00 } & \multicolumn{1}{l}{2.06 }  \\\cline{2-4}
                                      & \multicolumn{1}{l}{LiveSketch \cite{LS2019}} & \multicolumn{1}{l}{{ 25.8  }} & \multicolumn{1}{l}{30.9  }  \\\cline{2-4}
                                      & \multicolumn{1}{l}{TForm-Cont} & \multicolumn{1}{l}{{ 14.0  }} & \multicolumn{1}{l}{6.18  }  \\\cline{2-4}
                                      & \multicolumn{1}{l}{TForm-Tok-Grid} & \multicolumn{1}{l}{{ 19.4  }} & \multicolumn{1}{l}{17.5  }  \\\cline{2-4} 
                                      & \multicolumn{1}{l}{TForm-Tok-Dict} & \multicolumn{1}{l}{\bf 31.2  } & \multicolumn{1}{l}{{\bf 33.0  }}  \\\cline{2-4}
                                      & \multicolumn{1}{l}{{\em Uncertain}} & \multicolumn{1}{l}{8.60  } & \multicolumn{1}{l}{{ 10.3  }}  \\\hline
                                                                            
            \multirow{5}{*}{Mean ($50^\mathrm{th}$ cent.)}  & \multicolumn{1}{l}{SketchRNN \cite{Eck2018}} & \multicolumn{1}{l}{0.00  } & \multicolumn{1}{l}{0.00  }  \\\cline{2-4}
                                      & \multicolumn{1}{l}{LiveSketch \cite{LS2019}} & \multicolumn{1}{l}{ 25.2  } & \multicolumn{1}{l}{20.2  }  \\\cline{2-4}
                                      & \multicolumn{1}{l}{TForm-Cont} & \multicolumn{1}{l}{{ 15.6  }} & \multicolumn{1}{l}{16.0  }  \\\cline{2-4}
                                      & \multicolumn{1}{l}{TForm-Tok-Grid} & \multicolumn{1}{l}{{ 15.6  }} & \multicolumn{1}{l}{19.1  }  \\\cline{2-4} 
                                      & \multicolumn{1}{l}{TForm-Tok-Dict} & \multicolumn{1}{l}{\bf 35.8  } & \multicolumn{1}{l}{{\bf 35.1  }}\\\cline{2-4}
                                      & \multicolumn{1}{l}{{\em Uncertain}} & \multicolumn{1}{l}{7.36  } & \multicolumn{1}{l}{{ 9.57  }}\\\hline                                      

            \multirow{5}{*}{Long ($90^\mathrm{th}$ cent.)}  & \multicolumn{1}{l}{SketchRNN \cite{Eck2018}} & \multicolumn{1}{l}{0.00  } & \multicolumn{1}{l}{0.00  }  \\\cline{2-4}
                                      & \multicolumn{1}{l}{LiveSketch \cite{LS2019}} & \multicolumn{1}{l}{ 25.0  } & \multicolumn{1}{l}{21.1  }  \\\cline{2-4}
                                      & \multicolumn{1}{l}{TForm-Cont} & \multicolumn{1}{l}{ 12.5  } & \multicolumn{1}{l}{8.42  }  \\\cline{2-4}
                                      & \multicolumn{1}{l}{TForm-Tok-Grid} & \multicolumn{1}{l}{16.7  } & \multicolumn{1}{l}{10.5  }  \\\cline{2-4} 
                                      & \multicolumn{1}{l}{TForm-Tok-Dict} & \multicolumn{1}{l}{{\bf 40.6}  } & \multicolumn{1}{l}{{\bf 50.5  }}\\  \cline{2-4}
                                      & \multicolumn{1}{l}{{\em Uncertain}} & \multicolumn{1}{l}{5.21  } & \multicolumn{1}{l}{9.47  }\\\hline
        \end{tabular}

    \caption{User study quantifying interpolation quality for a pair of sketches of  same  (intra-) or between  (inter-)  classes.   Preference is expressed  by 5 independent workers, and results with $>50\%$ agreement are included.  Experiment repeated for short, medium and longer stroke sequences. }
    \label{tab:res_interp}
\end{table}
\subsection{Cross-modal Matching}

We evaluate the performance of Sketchformer for sketch based retrieval of sketches (S-S) and images (S-I).

\begin{figure*} [t!]
    \centering
    \begin{tabular}{c|c}
        \includegraphics[width=0.05\linewidth]{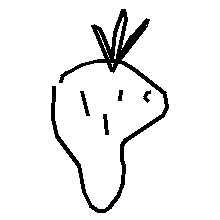}
         & \includegraphics[width=0.9\linewidth]{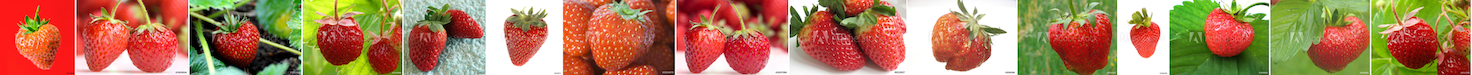}\\ 
         \includegraphics[width=0.05\linewidth]{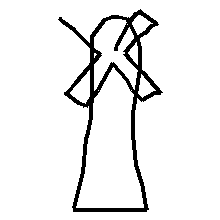}
         & \includegraphics[width=0.9\linewidth]{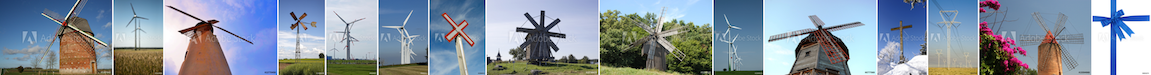}\\ 
         \includegraphics[width=0.05\linewidth]{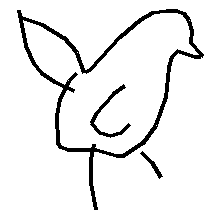}
         & \includegraphics[width=0.9\linewidth]{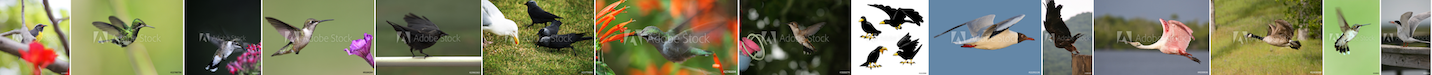} \\
         \includegraphics[width=0.05\linewidth]{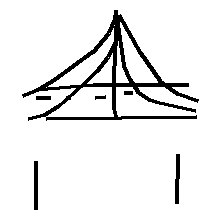}
         & \includegraphics[width=0.9\linewidth]{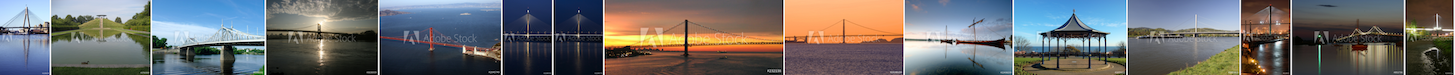} \\
         \includegraphics[width=0.05\linewidth]{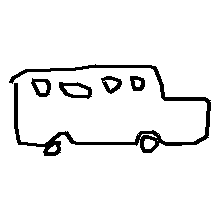}
         & \includegraphics[width=0.9\linewidth]{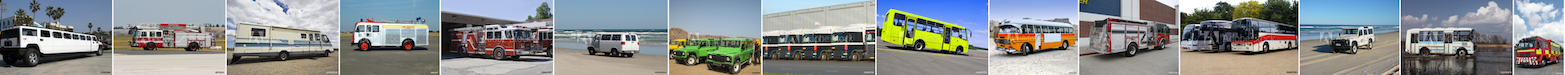} \\
         \includegraphics[width=0.05\linewidth]{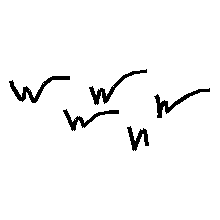}
         & \includegraphics[width=0.9\linewidth]{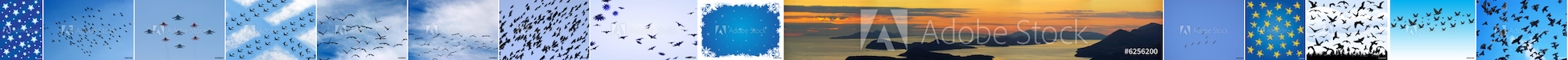}\\ 
         \includegraphics[width=0.05\linewidth]{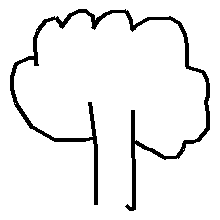}
         & \includegraphics[width=0.9\linewidth]{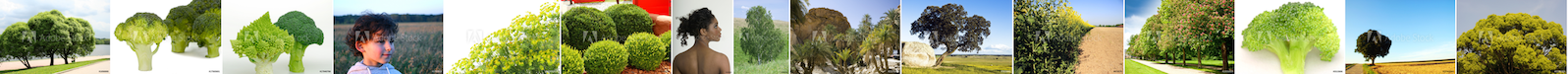}\\ 
    \end{tabular}
    \caption{Representative visual search results over Stock10M indexed by our proposed embedding (TForm-Tok-Dict) for a vector sketch query. The two bottom rows (`animal migration' and `tree') are failure cases.}
    \label{fig:sbir_vis}
\end{figure*}

\begin{figure}[t!]
    \centering
 \includegraphics[width=0.9\linewidth]{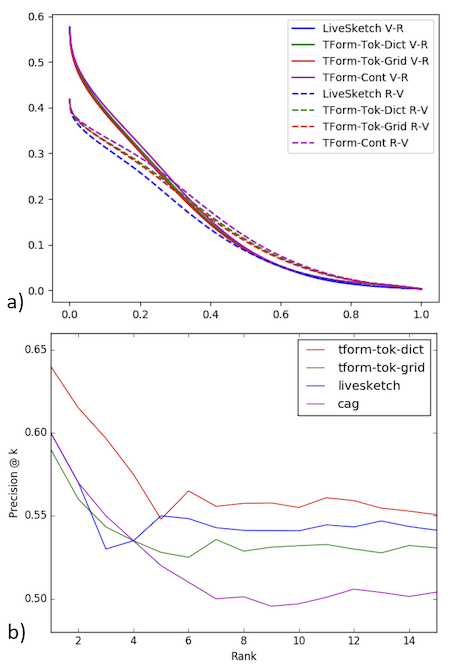}
    \caption{Quantifying search accuracy. a) Sketch2Sketch via precision-recall (P-R) curves for Vector-2-Raster and Raster-2-Vector category-level retrieval. b) Sketch2Image (SBIR) accuracy via precision @ k=[1,15] curve over Stock10M.}
    \label{fig:stockpk}
\end{figure}

\textbf{Sketch2Sketch (S-S) Matching.} We quantify the accuracy of retrieving sketches in one modality (raster) given a sketched query in another (vector, \ie stroke sequence) -- and vice-versa.  This evaluates the performance of Sketchformer in discriminating between sketched visual structures invariant to their input modality.  Sketchformer is trained on QD-2.5M and we query the test corpus QD-826k using QD-345Q as the query set.   We measure overall mean average precision (mAP) for both coarse grain (\ie class-specific) and fine-grain (\ie instance-level) similarity, as mean average of mAP for each query.  As per \cite{LS2019} for the former we consider a retrieved record a match if it matches the sketched object class.  For the latter,  exactly the same single sketch must match (in its different modality). To run raster variants, a rasterized version of QD-862k (for V-R) and of QD345-Q (for R-V) is produced by rendering strokes to a $256 \times 256$ pixel canvas using the CairoSVG Python library. Table~\ref{tab:res_mAP_SBSR}  show that for both class and instance level retrieval, the R-V configuration outperforms V-R indicating a performance gain due to encoding this large search index using the vector representation.  In contrast to other experiments reported, the continuous variant of Sketchformer appears slightly preferred, matching higher for early ranked results for the S-S case -- see Fig.~\ref{fig:stockpk}a for category-level precision-recall curve. Although Transformer outperforms RNN baselines by 1-3\% in the V-R case the gain is more limited and indeed the performance over baselines is equivocal in the S-S where the search index is formed of rasterized sketches.\\

\textbf{Sketch2Image (S-I) Matching.} 
We evaluate sketch based image retrieval (SBIR) over Stock10M dataset of diverse photos and artworks, as such data is commonly indexed for large-scale SBIR evaluation \cite{sketchstyle2017,LS2019}.  We compare against the state of the art SBIR algorithms accepting vector (LiveSketch \cite{LS2019}) and raster (Bui \etal \cite{Bui2018}) sketched queries. Since no ground-truth annotation is possible for this size of corpus, we crowd-source per-query annotation via Mechanical Turk (MTurk) for the top-$k$ ($k$=15) results and compute both mAP\% and precision@$k$ curve averaged across all QD345-Q query sketches.  Table~\ref{tab:res_mAP_SBIR} compares performance of our tokenized variants to these baselines, alongside associated Precision@k curves in  Fig.~\ref{fig:stockpk}b.  The proposed dictionary learned transformer embedding (TForm-Tok-Dict) delivers the best performance (visual results in  Fig.~\ref{fig:sbir_vis}).

\begin{table}[t!]
    \centering
        \begin{tabular}{llcc}
            \ttfamily Method & \ttfamily ~~ & \ttfamily Instance & \ttfamily Category \\                \hline
            \hline
            \multirow{2}{*}{Livesketch~\cite{LS2019}} & \multicolumn{1}{l}{V-R} & \multicolumn{1}{l}{6.71} & \multicolumn{1}{l}{20.49}
            \\
            & \multicolumn{1}{l}{R-V} & \multicolumn{1}{l}{7.15} & \multicolumn{1}{l}{20.93}
            \\\hline
            \multirow{2}{*}{TForm-Cont} & 
                \multicolumn{1}{l}{V-R} & \multicolumn{1}{l}{5.29} & \multicolumn{1}{l}{22.22}\\
                & \multicolumn{1}{l}{R-V} & \multicolumn{1}{l}{6.21} & \multicolumn{1}{l}{\textbf{23.48}}\\
            \hline
            \multirow{2}{*}{TForm-Tok-Grid} & 
                \multicolumn{1}{l}{V-R} & \multicolumn{1}{l}{6.42} & \multicolumn{1}{l}{21.26}\\
                & \multicolumn{1}{l}{R-V} & \multicolumn{1}{l}{\textbf{7.38}} & \multicolumn{1}{l}{22.10}\\
            \hline
            \multirow{2}{*}{TForm-Tok-Dict} & 
                \multicolumn{1}{l}{V-R} & \multicolumn{1}{l}{6.07} & \multicolumn{1}{l}{21.56}\\
                & \multicolumn{1}{l}{R-V} & \multicolumn{1}{l}{7.08} & \multicolumn{1}{l}{22.51}\\
            \hline
        \end{tabular}

    \caption{Quantifying the performance of Sketch2Sketch retrieval under two RNN baselines and three proposed variants.  We report category- and instance-level retrieval (mAP\%).}
    \label{tab:res_mAP_SBSR}
\end{table}

\begin{table}[t!]
    \centering
        \begin{tabular}{lll}
            \ttfamily Method & \ttfamily ~~ & \ttfamily mAP\% \\                \hline
            \hline
            \multirow{2}{*}{Baseline} & \multicolumn{1}{l}{LiveSketch \cite{LS2019}} & \multicolumn{1}{l}{54.80} \\\cline{2-3}
                                      & \multicolumn{1}{l}{CAG \cite{Bui2017}} & \multicolumn{1}{l}{51.97} \\\hline
            \multirow{3}{*}{Proposed} & \multicolumn{1}{l}{TForm-Tok-Grid} & \multicolumn{1}{l}{53.75} \\\cline{2-3}
                                    & \multicolumn{1}{l}{TForm-Tok-Dict} & \multicolumn{1}{l}{{\bf 56.96}} \\\hline
        \end{tabular}

    \caption{Quantifying accuracy of Sketchformer for Sketch2Image search (SBIR).  Mean average precision (mAP) computed to rank 15 over Stock10M for the QD345-Q query set.}
    \label{tab:res_mAP_SBIR}
\end{table}

%

\section{Conclusion}

We presented Sketchformer; a learned representation for sketches based on the Transformer architecture \cite{Vaswani2017}.  Several variants were explored using continuous and tokenized input; a dictionary learning based tokenization scheme delivers performance gains of 6\% on previous LSTM  autoencoder models (SketchRNN and derivatives). We showed interpolation within the embedding yields plausible blending of sketches within and between classes, and that reconstruction (auto-encoding) of sketches is also improved for  complex sketches.  Sketchformer was also shown effective as a basis for indexing sketch and image collections for sketch based visual search.  Future work could further explore our continuous representation variant, or other variants with more symmetric encoder-decoder structure.  We have demonstrated the potential for Transformer networks to learn a multi-purpose representation for sketch, but believe many further applications of Sketchformer exist beyond the three tasks studied here. For example, fusion with additional modalities might enable sketch driven photo generation \cite{park2019gaugan} using complex sketches, or with a language embedding for novel sketch synthesis applications.


{\small
\bibliographystyle{ieee_fullname}
\bibliography{egbib}
}

\end{document}